\documentclass[letterpaper, 10 pt, journal, twoside]{ieeetran}

\IEEEoverridecommandlockouts 

\usepackage{cite}
\usepackage{tabularx}
\usepackage{multirow}
\usepackage{amsmath,amssymb,amsfonts}
\usepackage{graphicx}
\usepackage{textcomp}

\usepackage[table,xcdraw]{xcolor}
\usepackage[linesnumbered,ruled,lined]{algorithm2e}
\usepackage{booktabs}
\usepackage{xcolor}
\usepackage{authblk}

\usepackage{enumitem}
\usepackage{hyperref}
\usepackage{breqn}
\usepackage{algpseudocode}
\usepackage{caption}
\usepackage{subcaption}
\usepackage{tcolorbox}
\usepackage{array}
\usepackage{makecell}
\usepackage{cuted}
\usepackage{adjustbox}

\usepackage[english]{babel}
\usepackage{amsthm}

\definecolor{my_green}{RGB}{144, 238, 144}

\allowdisplaybreaks

\title{ MonoMPC: Monocular Vision Based Navigation with Learned Collision Model and Risk-Aware Model Predictive Control }
\author{Basant Sharma$^{1}$, Prajyot Jadhav$^{1}$, Pranjal Paul$^{2}$, K.Madhava Krishna$^{2}$,Arun Kumar Singh$^{1, 3}$%
\thanks{Manuscript received: July 24, 2025; Revised September 10, 2025 ; Accepted November 8, 2025.}
\thanks{This paper was recommended for publication by Editor Aniket Bera upon evaluation of the Associate Editor and Reviewers' comments.
This work was in part supported by grant PSG753 from Estonian Research Council,
project TEM-TA101 funded by European Union and Estonian Research Council and SekMO program (2021-2027.1.01.23-0419) co-funded by European Union.}
\thanks{$^{1}$:Institute of Technology, University of
Tartu, Tartu, Estonia.
$^{2}$ : International Institute of Information Technology-Hyderabad, India, $^{3}$: Estonian Aviation Academy}%
\thanks{{\tt\footnotesize Emails: \{aks1812,basantsharma1990\}@gmail.com.}}
\thanks{{\tt\footnotesize Code: \url{https://github.com/Basant1861/MonoMPC}
Website: \url{https://sites.google.com/view/monompc}}}
\thanks{Digital Object Identifier (DOI): see top of this page.}
}

\makeatletter
\let\@oldmaketitle\@maketitle
\renewcommand{\@maketitle}{\@oldmaketitle
\centering
\includegraphics[width=0.82\linewidth]{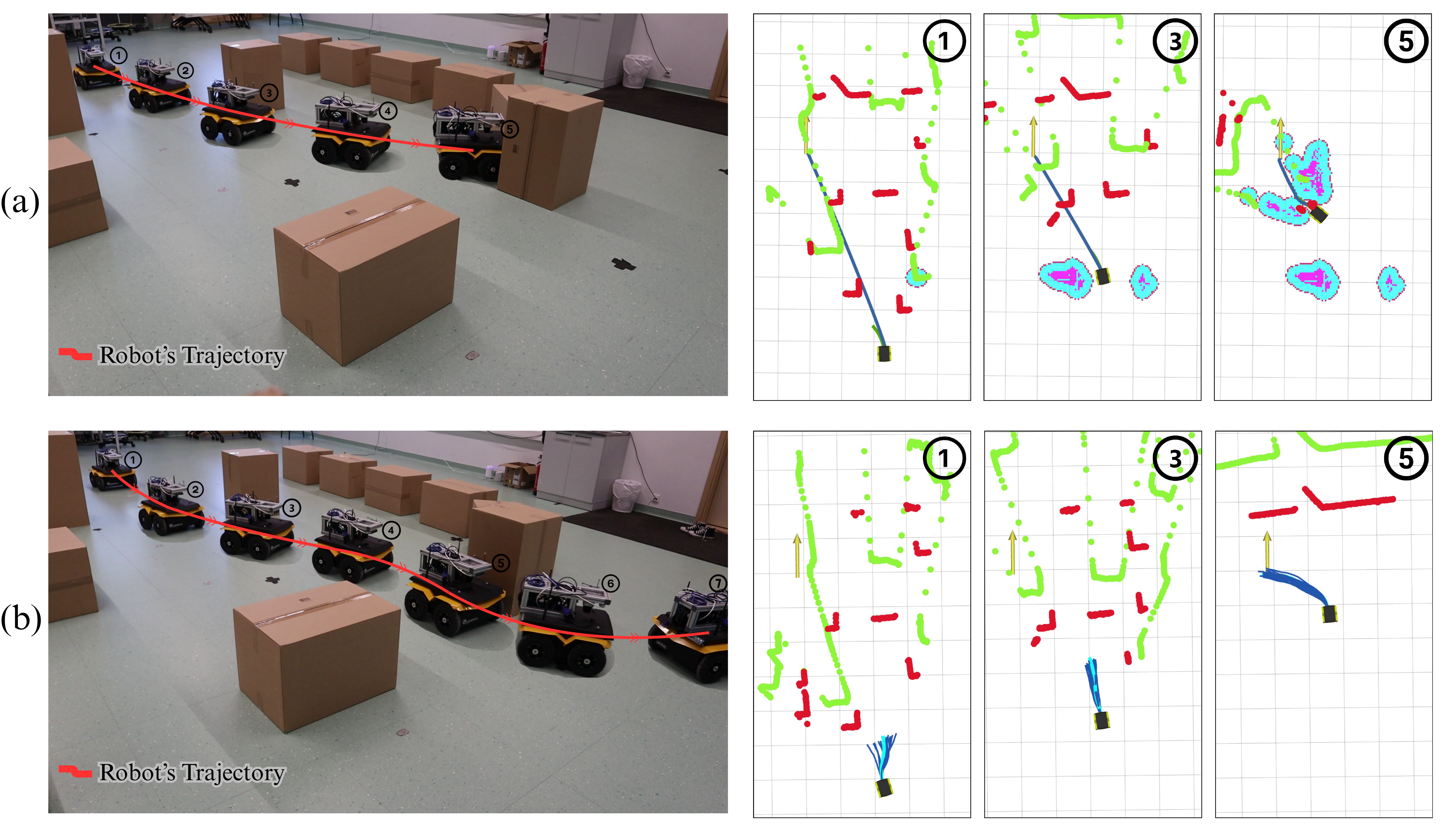}
\captionof{figure}{\footnotesize{Monocular navigation in cluttered environments using ROSNAV (top) vs. our approach (bottom). ROSNAV constructs cost maps directly from the estimated point cloud (green) generated by DepthAnything\cite{NEURIPS2024_26cfdcd8}, which deviates significantly from the ground-truth (red), leading to incorrect free-space detection (e.g., top row, panel 3) and collisions. In contrast, our method treats the estimated point cloud as a conditioning input to a learned probabilistic collision model, integrated with a risk-aware MPC framework. Snapshots across time steps are shown for both methods (corresponding time indices are labeled).}   }
\label{fig:teaser}
\vspace{-0.45cm}
}
\makeatother

\begin{document}

\thispagestyle{empty}

\begin{center}
\vspace*{0.05\textheight}

\begin{minipage}{0.9\textwidth}
\large{\copyright~2025 IEEE. Personal use of this material is permitted.
Permission from IEEE must be obtained for all other uses, in any current or future media, including
reprinting/republishing this material for advertising or promotional purposes,
creating new collective works, for resale or redistribution to servers or lists, or
reuse of any copyrighted component of this work in other works.}
\end{minipage}

\end{center}

\newpage

\setcounter{page}{1}

\maketitle




\begin{abstract}
Navigating unknown environments with a single RGB camera is challenging, as the lack of depth information prevents reliable collision-checking. While some methods use estimated depth to build collision maps, we found that depth estimates from vision foundation models are too noisy for zero-shot navigation in cluttered environments.
We propose an alternative approach: instead of using noisy estimated depth for direct collision-checking, we use it as a rich context input to a learned collision model. This model predicts the distribution of minimum obstacle clearance that the robot can expect for a given control sequence. At inference, these predictions inform a risk-aware MPC planner that minimizes estimated collision risk. We proposed a joint learning pipeline that co-trains the collision model and risk metric using both safe and unsafe trajectories. Crucially, our joint-training ensures well calibrated uncertainty in our collision model that improves navigation in highly cluttered environments. Consequently, real-world experiments show reductions in collision-rate and improvements in goal reaching and speed over several strong baselines. 

\end{abstract}


\section{Introduction}
\IEEEPARstart{N}{avigation} in unknown environments using a monocular RGB camera is well-suited for lightweight robotic platforms like aerial or compact ground robots, where size, weight, and power constraints make LiDARs impractical. Vision-only setups reduce hardware complexity and energy usage, supporting longer missions and broader deployment in cost-sensitive settings. However, monocular navigation remains fundamentally challenging due to the absence of direct depth perception, which hampers reliable collision-checking essential for safe operation in cluttered environments.

Recent progress in vision foundation models like DepthAnything~\cite{depthanything,NEURIPS2024_26cfdcd8} and ZoeDepth~\cite{https://doi.org/10.48550/arxiv.2302.12288} has enabled depth prediction from a single RGB image, opening new avenues for monocular navigation. Prior work has used these depth estimates to build local 3D or collision maps for planning~\cite{simon2023mononav}. However, depth estimates from existing models \cite{depthanything,NEURIPS2024_26cfdcd8}, \cite{https://doi.org/10.48550/arxiv.2302.12288}  are often too noisy or inconsistent for reliable collision-checking, particularly in close-proximity scenarios where small errors can result in collisions (see Fig.~\ref{fig:teaser}). In zero-shot settings, where the robot must generalize to unseen environments, these models fall short as standalone solutions and we further discuss these observation in Section~\ref{validation}.

\noindent \textbf{Contributions:} This work introduces a principled framework for autonomous navigation that, for the first time, directly reasons about uncertainty from estimated depth. At the core of our approach is a learned collision model that predicts a distribution over the minimum obstacle clearance for a planned trajectory. By conditioning this model on depth estimates, we can quantify collision risk at inference time. This risk is then integrated into a risk-aware Model Predictive Control (MPC) planner, which optimizes for actions that are both safe and goal-oriented.


A core innovation of our approach lies in appropriately transferring the noise in depth estimation to the actual downstream collision risk estimate. This is achieved by developing a joint learning pipeline  that co-trains the collision model and the risk metric using both safe and un-safe trajectories. Our approach ensures that during the training process, the probabilistic collision model is aware of how its predictions impact downstream risk. This in turn, provides crucial supervision signal during training that regularizes the variance of the collision model and enables safe yet non-conservative behavior.

We validate our approach on real-world hardware, showing significant navigation gains over two baseline groups. The first, including the ROS stack and MonoNav \cite{simon2023mononav}, builds occupancy maps from noisy depth estimates, often causing unsafe or overly conservative behavior (Fig.~\ref{fig:teaser}). The second, NoMaD \cite{10610665}-a diffusion-based end-to-end policy that directly regresses robot trajectory based on the input RGB image-lacks the precision needed for cluttered scenes. We also present additional statistical results validating the accuracy and statistical consistency of our learned collision model.

\begin{figure*}[t!]
     \centering
     \includegraphics[scale=0.55]{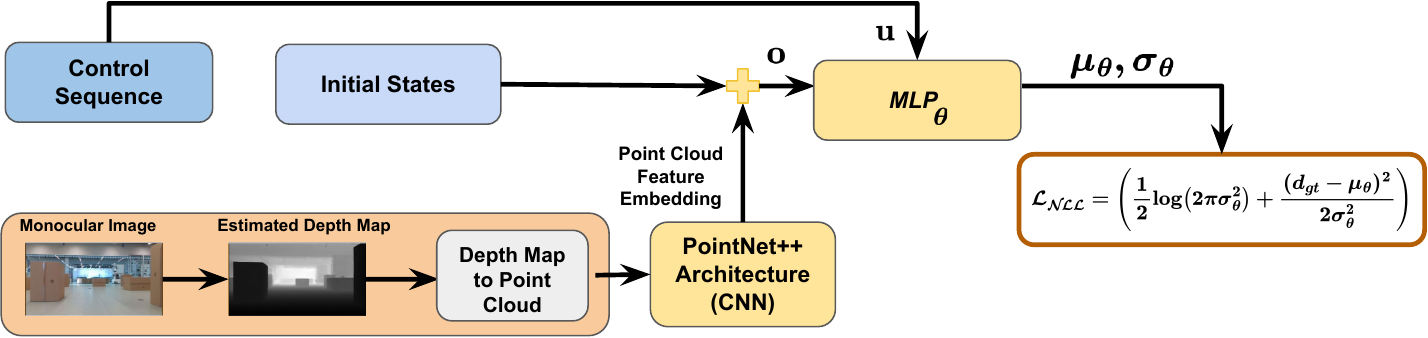}%
    \caption{\footnotesize{Overview of baseline learning pipeline for our probabilistic collision model that predicts worst-case obstacle clearance along a trajectory. Given an RGB image and control sequence, we extract geometric features from the estimated point cloud using a pre-trained depth estimator and PointNet++. Combined with the initial robot state, these form the observation vector, which an MLP uses to predict the mean and variance of obstacle clearance. The learnable components (yellow blocks) are trained end-to-end using Gaussian negative log-likelihood loss.}}
     \label{fig:vanilla_nll_pipeline}
\end{figure*}

\section{Related Works}
In the following, we restrict our review to works that use only monocular camera for navigation and exclude those that rely on depth sensors or LiDARs.

\noindent \textbf{End-to-End Approaches for Visual Navigation:} A common approach in monocular visual navigation is to train end-to-end models that map RGB images directly to control outputs like velocity commands or waypoints. Many rely on supervised learning via behavioral cloning or imitation learning~\cite{8460487},~\cite{8264734},~\cite{chen2020learning},~\cite{chaplot2020learning},~\cite{chaplot2020neural},~\cite{Sadeghi-RSS-17}. While simple and appealing, these methods often generalize poorly, are sensitive to perception noise, and lack interpretability. They also omit explicit modeling of geometry, uncertainty, and constraints, making them data-hungry and hard to adapt. Recent methods like ViNT~\cite{shah2023vint} and NoMaD~\cite{10610665} improve generalization by using image-conditioned diffusion models to generate trajectories from RGB input. However, they depend on high-quality supervision and do not model collision uncertainty.

Another class of approaches focuses on learning spatial representations from monocular input~\cite{chaplot2020learning,chaplot2020neural,simon2023mononav}. For example,~\cite{simon2023mononav} constructs a map from estimated depth on-the-fly and then performs planning over it. However, Section~\ref{validation} shows that errors in occupancy predictions due to noise in estimated depth proves detrimental in cluttered environments.

Deep reinforcement learning (DRL) has also been explored for visual navigation, particularly in unseen or partially known environments~\cite{7989381},~\cite{8460655},~\cite{10629078}. While DRL enables flexible end-to-end learning from experience, it typically requires extensive training in simulation, careful reward shaping, and offers limited safety guarantees.

\noindent \textbf{Predictive Models for Visual Navigation:} Approaches like ~\cite{Bar_2025_CVPR},~\cite{hansen2024tdmpc2},~\cite{wu2022daydreamer} learn action and image conditioned predictive models to simulate visual dynamics and leverage them for downstream planning and control, rather than directly learning policies. ~\cite{8750823} adopt a slightly different approach and propose a visual MPC controller that outputs a sequence of velocity commands based on the current image and a trajectory of subgoal images. Authors in \cite{jacquet2024n} regress depth images and actions to collision probabilities which is then minimized within an MPC framework. Although theoretically, this approach can be extended to work with estimated depth, it is unclear how the associated noise will affect the efficiency of the overall pipeline. Moreover, we believe that predicting obstacle clearance distribution is easier than directly predicting collision probabilities, as the supervision data for the former can be easily obtained. Our design choice also allows us to use sophisticated statistical tools to capture collision risk from the clearance samples in an efficient manner.

\noindent \textbf{Improvement Over State-of-the-Art:} Unlike end-to-end pipelines, our approach explicitly incorporates safety constraints by learning an action-conditioned collision model, placing us closer to ~\cite{Bar_2025_CVPR},~\cite{hansen2024tdmpc2},~\cite{wu2022daydreamer},  \cite{8750823}.However, in contrast to these cited works, we explicitly reason about uncertainty in collision prediction, and introduce a novel method to learn task-aware uncertainty. Moreover, we also learn optimal parameters of our risk metric to improve the efficacy of our approach. 
On the implementation side, most monocular navigation methods demonstrate results in simple environments like hallways. In contrast, we show, for the first time, robust and reliable navigation in cluttered settings typically requiring LiDAR or depth sensors.

\section{Main Algorithmic Results}
\subsubsection*{Symbols and Notations} We use small/upper case normal-font to represent scalars. The bold-face small fonts represent vectors while upper-case variants represent matrices.

\subsection{Vision-Based Navigation as Risk-Aware Trajectory Optimization}
\noindent We frame monocular vision-based collision avoidance and goal reaching as the following trajectory optimization.

\vspace{-0.3cm}
\small
\begin{subequations}
\begin{align}
    \min_{\mathbf{u}} w_1c(\mathbf{x} )+ w_2r(d(\mathbf{u},\mathbf{o}))+ w_3\left\Vert \begin{matrix}
        \mathbf{u}
    \end{matrix}\right\Vert_2^2 \label{cost},\\
    \mathbf{x}_{k+1} = f(\mathbf{x}_k, \mathbf{u}_k), \ \mathbf{u}_{min}\leq \mathbf{u}_k\leq \mathbf{u}_{max} \forall k\label{control_con}
\end{align}
\end{subequations}
\normalsize

\noindent where $c(\cdot)$ represents the state-dependent cost function. The vector $\mathbf{x}_k$ represents the state of the robot at time-step $k$. The vector $\mathbf{x}$ is the concatenation of the states at different $k$. The function $f$ represents the robot dynamics model. The control inputs are represented by $\mathbf{u}_k$. The vector $\mathbf{u}$ is formed by stacking $\mathbf{u}_k$ at different time-step $k$ respectively. The scalar function $d$ represents a learned probabilistic collision model, which takes as input the observations $\mathbf{o}$ and the control sequence $\mathbf{u}$ and outputs a distribution of worst-case obstacle clearance along the trajectory resulting from $\mathbf{u}$. We discuss the details of $d$ in the subsequent sections. The first term $c$ in \eqref{cost} minimizes the state cost, typically addressing path-following errors. The second term $r$ captures collision-risk based on the predictions of $d$. The last term $\left\Vert \mathbf{u} \right\Vert_2^2$ in \eqref{cost} penalizes large control inputs, and the weights $w_i$ tune the robot's risk-seeking behavior. Control bounds are enforced through \eqref{control_con}. We construct a MPC feedback loop by solving \eqref{cost}-\eqref{control_con} in a receding horizon manner from the current state.

\begin{figure*}[t!]
    \centering
    \includegraphics[scale=0.58]{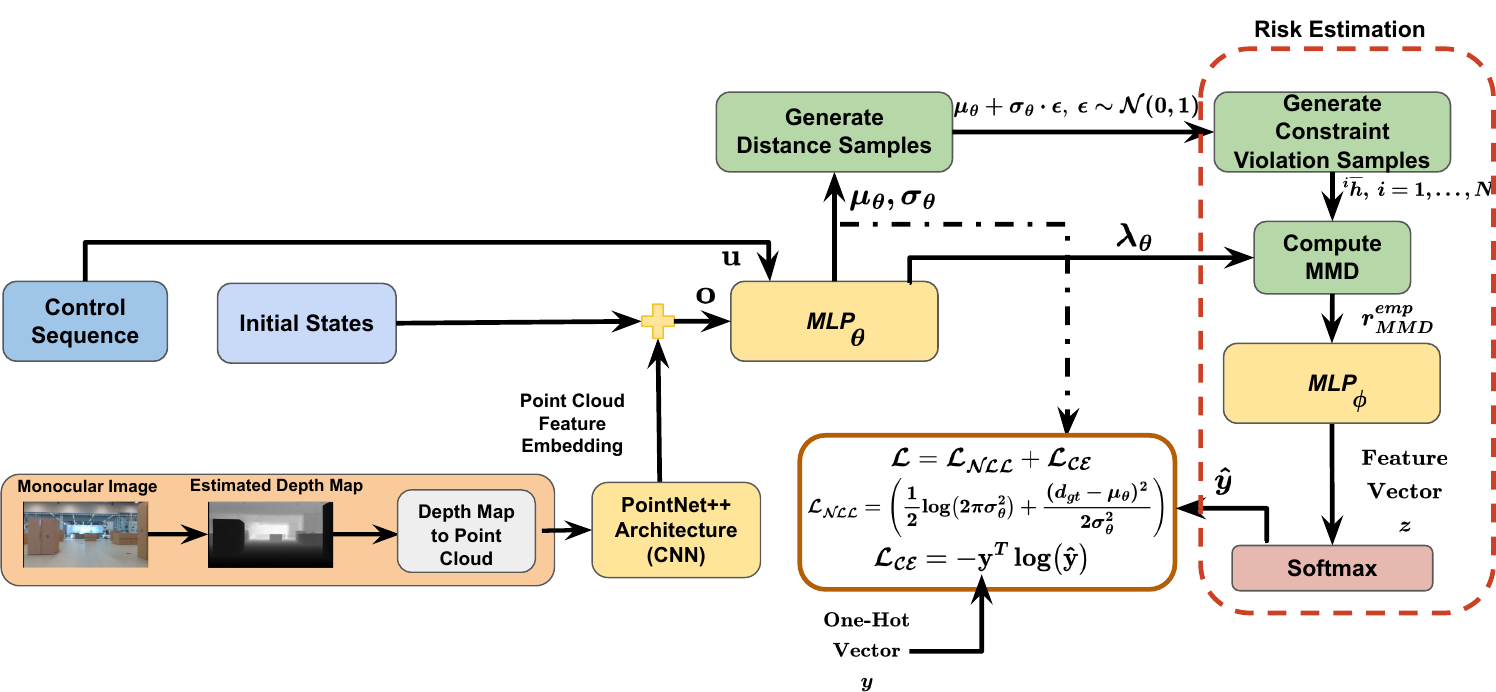}%
    \caption{\footnotesize{Overview of our task-aware learning of probabilistic collision model. The previous baseline model of Fig.~\ref{fig:vanilla_nll_pipeline} had a weak supervision on predicted variance due to the absence of ground-truth uncertainty, often resulting in over- or underconfident predictions. To address this, we introduce downstream supervision via collision risk estimation. The observation vector and control sequence are passed through $\text{MLP}_\theta$ to predict the mean, variance, and a kernel parameter. Using the reparameterization trick, we generate obstacle clearance samples to compute constraint violations, forming an MMD-based risk representation. This is processed by $\text{MLP}_\phi$, followed by a softmax layer. The learnable parts shown in yellow are trained end-to-end with Gaussian NLL and a cross-entropy loss.}}
     \label{fig:aug_nll_pipeline}
\end{figure*}

\vspace{-0.4cm}
\subsection{Modeling Risk Through Chance-Constraints}
\noindent We model the robot's footprint as a circular disk of radius $d_o$. Thus, we can define $h(d(\mathbf{u}, \mathbf{o})) = -d(\mathbf{u}, \mathbf{o})+d_o \leq 0$ as the control and observation dependent collision-avoidance constraint. We define risk $r$ in terms of so-called chance constraints. These have the general form of $P(h(d(\mathbf{u}, \mathbf{o}))\geq 0)\leq \varepsilon$, where $P$ represents probability and $\varepsilon$ is some constant. Thus, we define risk as:
\begin{align}
    r(h(d(\mathbf{u}, \mathbf{o}))) = P(h(d(\mathbf{u}, \mathbf{o}))\geq 0)
    \label{risk_def}
\end{align}

\noindent Intuitively, our risk model captures the probability of collision constraints being violated for a given control sequence $\mathbf{u}$ and current observation $\mathbf{o}$. Moreover, the risk depends on the distribution characterized by the collision model $d$. Our aim is to first learn a suitable $d$ and then minimize the associated risk using \eqref{cost}-\eqref{control_con}.

A key challenge in using risk $r$ \eqref{risk_def} is that the left-hand side does not have an analytical form if the predictive distribution of  $d(\mathbf{u}, \mathbf{o})$ departs significantly from Gaussian. Thus, in the next subsection, we present a tractable surrogate risk model which can produce the same effect as $r$.

\subsection{Maximum Mean Discrepancy (MMD) as parameterized risk metric}
\noindent Let us define a constraint residual function as 
\begin{align}
    \overline{h}(d(\mathbf{u}, \mathbf{o})) = \max(0, h(d(\mathbf{u}, \mathbf{o}))
    \label{h_bar}
\end{align}
\noindent In the deterministic scenario, driving $\overline{h}(d(\mathbf{u}, \mathbf{o}))$ to zero will push $h(d(\mathbf{u}, \mathbf{o}))$ to the feasible boundary. In the stochastic case, \eqref{h_bar} maps $d(\mathbf{u}, \mathbf{o})$ to a distribution of constraint residuals. Let $\overline{h}(d(\mathbf{u}, \mathbf{o})) \sim p_{\overline{h}}$. Although we don't know the parametric form for $p_{\overline{h}}$, we can be certain that its entire mass lies to the right of $\overline{h} = 0$. Moreover, as $P(h(d(\mathbf{u}, \mathbf{o}))\geq 0)$ approaches zero, $p_{\overline{h}}$ converges to a Dirac-Delta distribution $p_{\delta}$ \cite{10979910}. In other words, one way of reducing risk is to minimize the difference between $p_{\overline{h}}$ and $p_{\delta}$. We formulate this distribution matching through the lens of Maximum Mean Discrepancy (MMD) \cite{10.5555/2188385.2188410},\cite{10979910} and use the following surrogate for the collision-risk proposed in  \cite{10979910}, \cite{11084873}.
\begin{align}
    r \approx r_{MMD}^{emp} = \left\Vert \hat{\mu}[\overline{h}]-\hat{\mu}[\delta] \right\Vert_{\mathcal{H}}^2,
    \label{risk_mmd_emp}
\end{align}

\noindent where, $\hat{\mu}[\overline{h}]$ is called the empirical RKHS embedding of $p_{\overline{h}}$. It is computed through the following expression
\begin{align}
    \hat{\mu}[\overline{h}] = \sum_{i = 1}^{i = N}\frac{1}{N}K_{\lambda} ({^{i}}\overline{h}, .),\label{hbar_emb_emp}
\end{align}    
\noindent where, ${^{i}}\overline{h} = \overline{h}({^{i}}d(\mathbf{u}, \mathbf{o}))$ and ${^i}d$ are the samples drawn from the predictive distribution characterized by $d$. In a similar manner, $\hat{\mu}[\delta]$ is computed based on the $N$ samples of $\delta$ drawn from $p_{\delta}$\footnote{We can approximate $p_{\delta}$ through a Gaussian $\mathcal{N}(0, \epsilon)$, with an extremely small covariance $\epsilon$ ($\approx 10^{-5}$).}. For any given $N$, as $r_{MMD}^{emp}\rightarrow 0$, $p_{\overline{h}} \rightarrow p_{\delta}$. Thus our aim is to compute a control sequence $\mathbf{u}$ which leads to as low as possible $r_{MMD}^{emp}$.

The $K_{\lambda}$ is a positive definite function in RKHS known as the kernel function with the "reproducing" property \eqref{repr_prop_kernel} \cite{10.5555/2188385.2188410} , where $\left\langle \cdot \right\rangle$ represents the inner product. Throughout this paper, we have used Laplacian kernel for which $\lambda$ represents the kernel width.
\begin{align}
    K_{\lambda}(\mathbf{z},\mathbf{z'}) = \left\langle K_{\lambda}(\mathbf{z},\cdot), K_{\lambda}(\mathbf{z'},\cdot)\right\rangle_{\mathcal{H}} \label{repr_prop_kernel}
\end{align}

\subsubsection*{Importance of Kernel Parameter} The kernel parameter \(\lambda\) critically controls the MMD metric, \(r_{\text{MMD}}^{\text{emp}}\), by shaping the embeddings \(\hat{\mu}[\overline{h}]\) and \(\hat{\mu}[\delta]\). An overly large \(\lambda\) causes the embeddings to collapse to a constant, making them indistinguishable and insensitive to collision risk. Conversely, a \(\lambda\) that is too small makes the embeddings artificially distinct, leading to overly conservative behavior that flags open spaces as high-risk. To resolve this, our approach learns to predict an optimal \(\lambda\) directly from current observations and control sequences (Section~\ref{augmented_model}, Fig.~\ref{fig:aug_nll_pipeline}).

\vspace{-0.4cm}
\subsection{Learning the Collision Model: The Baseline Approach}\label{vanilla_model}
\noindent This section presents our approach for learning a probabilistic collision model \(d(\mathbf{u}, \mathbf{o})\) which quantifies the worst-case obstacle clearance along a given trajectory. We model it as a Gaussian random variable with mean $\mu_{\boldsymbol{\theta}}(\mathbf{u}, \mathbf{o})$ and variance $\sigma_{\boldsymbol{\theta}}(\mathbf{u}, \mathbf{o})$, both of which are designed as feed-forward neural networks parameterized by weight $\boldsymbol{\theta}$.Our design choice captures observation noise in $\mathbf{o}$, primarily arising from depth or point-cloud estimates of the vision foundation model \cite{chua2018deep}.
The detailed architecture is shown in Fig.~\ref{fig:vanilla_nll_pipeline}. It consists of a pre-trained depth estimator that takes in RGB image and produces a depth-map. The depth-map is converted to a 2D point cloud and passed through PointNet++ \cite{NIPS2017_d8bf84be} to get a feature embedding. We concatenate PointNet++ features with the initial state to get the full observation vector $\mathbf{o}$, which is then passed along with control sequence $\mathbf{u}$ to a Multi-Layer Perceptron (MLP) to get $\mu_{\boldsymbol{\theta}}(\mathbf{u}, \mathbf{o})$ and $\sigma_{\boldsymbol{\theta}}(\mathbf{u}, \mathbf{o})$. Both PointNet++ and MLP are trained in end-to-end manner using the following negative log-likelihood (NLL) function.
\begin{align}
    \mathcal{L}_{\mathcal{NLL}} = \left( \frac{1}{2} \log(2\pi\sigma_{\boldsymbol{\theta}}^2) + \frac{(d_{gt} - \mu_{\boldsymbol{\theta}})^2}{2\sigma_{\boldsymbol{\theta}}^2} \right),
    \label{nll_loss}
\end{align}

\noindent where $d_{gt}$ represents the ground-truth worst-case obstacle clearance along a given trajectory. We explain how these are obtained in Section~\ref{validation}.

\noindent 
\begin{algorithm*}[!t]
\caption{Sampling-Based Optimizer to Solve \eqref{cost}-\eqref{control_con}}
\label{algo_1}
\SetAlgoLined
$M$ = Maximum number of iterations\\
Initiate mean $^{m}\boldsymbol{\nu}, ^{m}\boldsymbol{\Sigma}$, at iteration $m=0$ for sampling control inputs $\mathbf{u}$; Given observation vector $\mathbf{o}$, trained neural network $\boldsymbol{\theta}$(Fig.~\ref{fig:aug_nll_pipeline}) \\
\For{$m=1, m \leq M, m++$}
{
     \vspace{0.1cm}
    Initialize $CostList$ = []\\
     \vspace{0.1cm}
    
    Draw batch of ${n}$ control sequences $(\mathbf{u}_{1}, \mathbf{u}_{2}, \mathbf{u}_{q}, ...., \mathbf{u}_{{n}})$ from $\mathcal{N}(^{m}\boldsymbol{\nu}, ^{m}\boldsymbol{\Sigma})$ \\
    
    Query collision model $\forall \mathbf{u}_{q}$: $(\mu_{\boldsymbol{\theta},q},\sigma_{\boldsymbol{\theta},q},\lambda_{\boldsymbol{\theta},q})$ = $\text{MLP}_{\boldsymbol{\theta}}(\mathbf{u}_q,\mathbf{o})$ \tcp*[f]{\textcolor{blue}{$\text{MLP}_{\boldsymbol{\theta}}$ denotes the learned model whose output $\mu_{\boldsymbol{\theta},q},\sigma_{\boldsymbol{\theta},q},\lambda_{\boldsymbol{\theta},q}$ denotes the predicted mean, predicted variance and the kernel parameter, respectively, for each control sequence $\mathbf{u}_q$}}\\
    \vspace{0.1cm}

    Compute $N$ distance samples $^{i}{d_{q}}\sim\mathcal{N}(\mu_{q},\sigma_{q}) ,  \forall i=1,\ldots,N$ and subsequently $N$ constraint violation samples $^{i}\overline{h}_q$. Repeat this process $\forall q = (1, 2, \dots, n)$ \tcp*[f]{\textcolor{blue}{$^{i}{d_{q}}$ denotes $i$\text{-th} obstacle clearance sample corresponding to the $q$\text{-th} control sequence. Similarly for $^{i}\overline{h}_q$ }}\\  
    \vspace{0.1cm}
    
    Compute $\hat\mu_q[^{i}\overline{h}_q]$ over the constraint violation samples $^{i}\overline{h}_q$ through \eqref{hbar_emb_emp} and subsequently $r_{MMD,q}^{emp}$. Repeat this $\forall q = (1, 2, \dots, n)$ \tcp*[f]{\textcolor{blue}{Compute the MMD-based surrogate for collision risk for each $q$}}\\
    
    \vspace{0.1cm}
    
    Generate state trajectories $\mathbf{x}_q(\mathbf{u}_{q}), \forall q = (1, 2, \dots, n)$ \\
    
    \vspace{0.1cm}
    
    $ConstraintEliteSet \gets$ Select top $n_{c}$ batch of $\mathbf{u}_{q}, \mathbf{x}_q$ with lowest $r_{MMD,q}^{emp}$ \tcp*[f]{\textcolor{blue}{We sort control sequences $\mathbf{u}_q$ and trajectories $\mathbf{x}_q$ by lowest $r_{MMD,q}^{emp}$ values.}}\\
    
    \vspace{0.1cm}
    
    Define $c_q = w_1c(\mathbf{x}_q ) + w_2r_{MMD,q}^{emp}+w_3\left\Vert \begin{matrix}
            \mathbf{u}_q
        \end{matrix}\right\Vert_2^2 $\\
    
    \vspace{0.1cm}

    $cost \gets$ $c_q$,  $\forall q$ in the $ConstraintEliteSet$ \\
    \vspace{0.1cm}
    
    append each computed ${cost}$ to $CostList$ \\
    \vspace{0.1cm}
        
    $EliteSet  \gets$ Select top $n_{e}$ samples of ($\mathbf{u}_{q}, \mathbf{x}_q $) with lowest cost from $CostList$.\\
     \vspace{0.1cm}
    $({^{m+1}}\boldsymbol{\nu}, {^{m+1}}\boldsymbol{\Sigma} ) \gets$ Update distribution based on $EliteSet$ 
}
\Return{ Control inputs $\mathbf{u}_{q}$ and  $\mathbf{x}_q$ corresponding to lowest cost in the $EliteSet$}
\normalsize
\end{algorithm*}

\vspace{-0.4cm}
\subsection{Learning the Collision Model: The Down-Stream Task-Aware Approach}\label{augmented_model}
\noindent The core limitation of training with just the baseline NLL loss \eqref{nll_loss} is that the supervision on variance is rather weak since there is no ground truth variance available. Moreover, the training process is unaware of how its predictions are leveraged by the downstream task. Depending on how the training proceeds, the trained model can be either under-confident (high-variance) or overconfident(low-variance). For example, consider a scene where the actual $d_{gt}$ signifies a collision-free scenario. That is $h(d_{gt}) = -d_{gt}+d_0\leq 0$. Now, imagine that for this situation, the network of Fig.~\ref{fig:vanilla_nll_pipeline} predicts a $\mu_{\boldsymbol{\theta}}(\mathbf{u}, \mathbf{o})$ which is very close to $d_{gt}$ but the variance is unreasonably higher. In such a case, the collision-free scenario will be incorrectly assigned high-risk due to uncertainty stemming from a higher variance.

The training process described in this section is designed to overcome the above described issues. Our main insight is that the supervision on variance can and should come from the downstream task of estimating collision risk based on the predicted $\sigma_{\boldsymbol{\theta}}$. To this end, we present the modified architecture in Fig.~\ref{fig:aug_nll_pipeline}, which is obtained by adding a risk estimation head to the baseline architecture of Fig.~\ref{fig:vanilla_nll_pipeline}. During training time, we use the predicted $\mu_{\boldsymbol{\theta}}(\mathbf{u}, \mathbf{o})$, $\sigma_{\boldsymbol{\theta}}(\mathbf{u}, \mathbf{o})$ to generate samples $^{i}{d}$ of the random variable $d(\mathbf{u}, \mathbf{o})$. We use the re-parameterization trick \eqref{rep_trick} to generate samples in a differentiable manner. The samples $^{i}{d}$ should characterize zero risk if the associated $d_{gt}$ is indeed collision-free. On the other hand, the samples should lead to a high risk value if $d_{gt}$ is insufficient for collision avoidance. We can interpret this requirement as a two class classification problem conditioned on the samples $^{i}{d}$ and implicitly on the predicted variance  $\sigma_{\boldsymbol{\theta}}(\mathbf{u}, \mathbf{o})$. The risk estimation head of Fig.~\ref{fig:aug_nll_pipeline} performs exactly this role.

The risk estimation head estimates collision constraint violations from $^{i}{d}$ (\eqref{violation_training}). Using these violations and the predicted kernel width $\lambda_{\boldsymbol{\theta}}$, the empirical MMD risk $r_{MMD}^{emp}$ is computed. This risk is then passed through an MLP to yield a 2D feature vector $\mathbf{z}$, which a softmax layer normalizes to $\hat{\mathbf{y}}$. The ground-truth label for $\hat{\mathbf{y}}$  is given by $\mathbf{y}$ and is computed through \eqref{one_hot}

\vspace{-0.3cm}
\small
\begin{subequations}
    \begin{align}
        ^{i}{d} = \mu_{\boldsymbol{\theta}} + \sigma_{\boldsymbol{\theta}}\cdot {^{i}}\epsilon ,\ {^{i}}\epsilon \sim \mathcal{N}(0,1) \ , \forall i=1,\ldots,N \label{rep_trick}\\
        ^{i}h(^{i}d) = -{^{i}}{d} + d_{o}, \ ^{i}\overline{h}(^{i}d) = \max(0, {^{i}}h(^{i}d)) \label{violation_training} \\
        \mathbf{y} = 
\begin{cases}
[0, 1]^T, & \text{if } -d_{gt} + d_o \leq 0 \\
[1, 0]^T, & \text{if } -d_{gt} + d_o > 0 \\
\end{cases} \label{one_hot}
\\
\mathcal{L} = \mathcal{L_{NLL}} + \mathcal{L_{CE}}
\ ,\mathcal{L_{CE}} = -\mathbf{y}^T \log(\hat{\mathbf{y}}) \label{cross_entr_loss}
    \end{align}
\end{subequations}
\normalsize

The architecture of Fig.~\ref{fig:aug_nll_pipeline} is trained with a combination of NLL and a cross-entropy loss \eqref{cross_entr_loss}. The latter acts as an implicit task-aware regularization on the predicted $\sigma_{\boldsymbol{\theta}}(\mathbf{u}, \mathbf{o})$. 

\subsection{Overall Approach}
\noindent Alg.~\ref{algo_1} presents our overall approach wherein we solve \eqref{cost}-\eqref{control_con} using the trained neural worst-case obstacle clearance model as our collision predictor. We adopt a sampling-based optimizer based on \cite{bhardwaj2022storm}, wherein a batch of control sequences are drawn from a distribution which is gradually refined across iterations.

The algorithm initializes the sampling distribution (Line 2) and samples control inputs (Line 5), which are passed through a trained neural network (Line 6) to predict the mean $\mu_{\boldsymbol{\theta},q}$ and variance $\sigma_{\boldsymbol{\theta},q}$ associated with the distribution of obstacle clearances, as well as the kernel hyperparameter $\lambda_{\boldsymbol{\theta},q}$. These predictions generate $N$ distance samples $^i d_q$, constraint violations $^i\overline{h}_q$ (Line 7), and subsequently $r_{MMD,q}^{emp}$ (Line 8). State trajectories $\mathbf{x}_q(\mathbf{u}_q)$ are then generated using the robot dynamics (Line 9). The $n_c$ lowest-risk samples form the $ConstraintElliteSet$ (Line 10), from which costs are evaluated and stored (Lines 11–13). Finally, the top $n_e$ samples form the $ElliteSet$ (Line 14) used to update the sampling distribution (Line 15) following \cite{bhardwaj2022storm}.


\section{Validation and Benchmarking}\label{validation}
\subsection{Implementation Details}
\noindent We implemented Alg.~\ref{algo_1} in Python using Jax \cite{jax2018github} as the GPU-accelerated linear algebra back-end. The state cost consists of a running goal cost and has the following form:
\begin{align}
    c(\mathbf{x}) = \sum_k (x_k-x_f)^2 +(y_k-y_f)^2
\end{align}
where $x_f,y_f$ denote the end-point coordinates. We used Alg.~\ref{algo_1} in a receding horizon manner from the current state to create a MPC feedback loop.

\noindent \textbf{Data Collection:} We collected two datasets: one via teleoperation and another using Alg.~\ref{algo_1}-first with ground-truth LiDAR for collision avoidance to train the initial collision model and subsequently using Alg.~\ref{algo_1} with the learned collision model to collect further refinement data. Data were recorded at 10 Hz, with each sample consisting of:
\begin{itemize}
    \item A standardized estimated point cloud of shape $300 \times 2$ obtained from Depth-Anything-V2~\cite{NEURIPS2024_26cfdcd8},
    \item Robot state $\mathbf{x} = (x, y, \psi, v, \omega)$, where $(x, y)$ is position, $\psi$ is heading, and $v$, $\omega$ are the linear and angular velocities (sampled uniformly from $\mathcal{U}[-1,1]$ and clipped to $[0, 1]$ m/s and $[-1, 1]$ rad/s, respectively),
    \item A control sequence $\mathbf{u} = (\mathbf{v}, \boldsymbol{\omega})$ where $\mathbf{v},\boldsymbol{\omega}$ are the concatenation of the linear and angular velocities, respectively, at different timesteps,
    \item Ground-truth clearance $d_{gt}$ from LiDAR (see~\ref{gt_dist_comp}). This is the privileged information used during training.
\end{itemize}

\noindent \textbf{Control Sequence Generation:} Each control sequence consists of 50 steps over a $T=5$\,s horizon with sampling time of $0.1$\,s. Control sequences $\mathbf{u} = (\mathbf{v}, \boldsymbol{\omega})$ are generated via a uniform distribution in $[-1.0, 1.0]$ with $\mathbf{v}$ clipped to $[0, 1]$ m/s. The robot state evolves via:
\begin{align}
    x_{k+1} = x_k + v_k \cos(\psi_k) \Delta t ,\ y_{k+1} = & y_k + v_k \sin(\psi_k) \Delta t \nonumber \\
    \psi_{k+1} = \psi_k + \omega_k \Delta t
\end{align}

\noindent \textbf{Ground-Truth Distance Computation} \label{gt_dist_comp}
For each trajectory resulting from the $j\text{-th}$ control sequence $\mathbf{u}_j = (\mathbf{v}_j, \boldsymbol{\omega}_j), \ ,\forall j=1,\ldots,N_p$, the minimum distance to the LiDAR point cloud $\mathcal{P}_{\text{LiDAR}}$ is:
\begin{equation}
    d_{gt,j} = \min_{k,\,\mathbf{p} \in \mathcal{P}_{\text{LiDAR}}} 
    \left\| 
        \begin{bmatrix}
            x_{k,j} \\ y_{k,j}
        \end{bmatrix} - \mathbf{p} 
    \right\|_2
\end{equation}

\noindent \textbf{Training Set-up:} The probabilistic collision model (Sec.~\ref{augmented_model}) is trained in Equinox \cite{kidger2021equinox} on an RTX 5090 desktop. Inputs include estimated depth, vehicle states, and control sequences. The outputs are the mean and variance of obstacle clearance distribution. We collected 80k point-cloud instances. In each of them, we used $N_p = 50$ control sequences with randomized robot initial state to generate trajectories along which we computed the ground-truth worst-case obstacle clearance. This yielded a total of $\approx 6$M training points(point-cloud times $N_p$ times random initial states). The total training time was $\approx$2.5\,h, following the procedure in Secs.~\ref{vanilla_model}–\ref{augmented_model}.

\noindent \textbf{Benchmarking Setup:} We evaluate our approach using a Clearpath Jackal robot equipped with an Intel RealSense D435i RGB-D camera (RGB-only input, 69° horizontal FOV). For visualization and analysis, LiDAR data was recorded from a SICK TIM551 2D sensor. The system ran on ROS Noetic with an NVIDIA RTX 3080 GPU laptop connected to the Jackal over a local network. Depth estimation was performed in real time using TensorRT-accelerated Depth-Anything-V2~\cite{NEURIPS2024_26cfdcd8}. Experiments were conducted in indoor environments with static cardboard boxes to simulate clutter.

\begin{figure}[!t]
    \centering
    \includegraphics[scale=0.14]{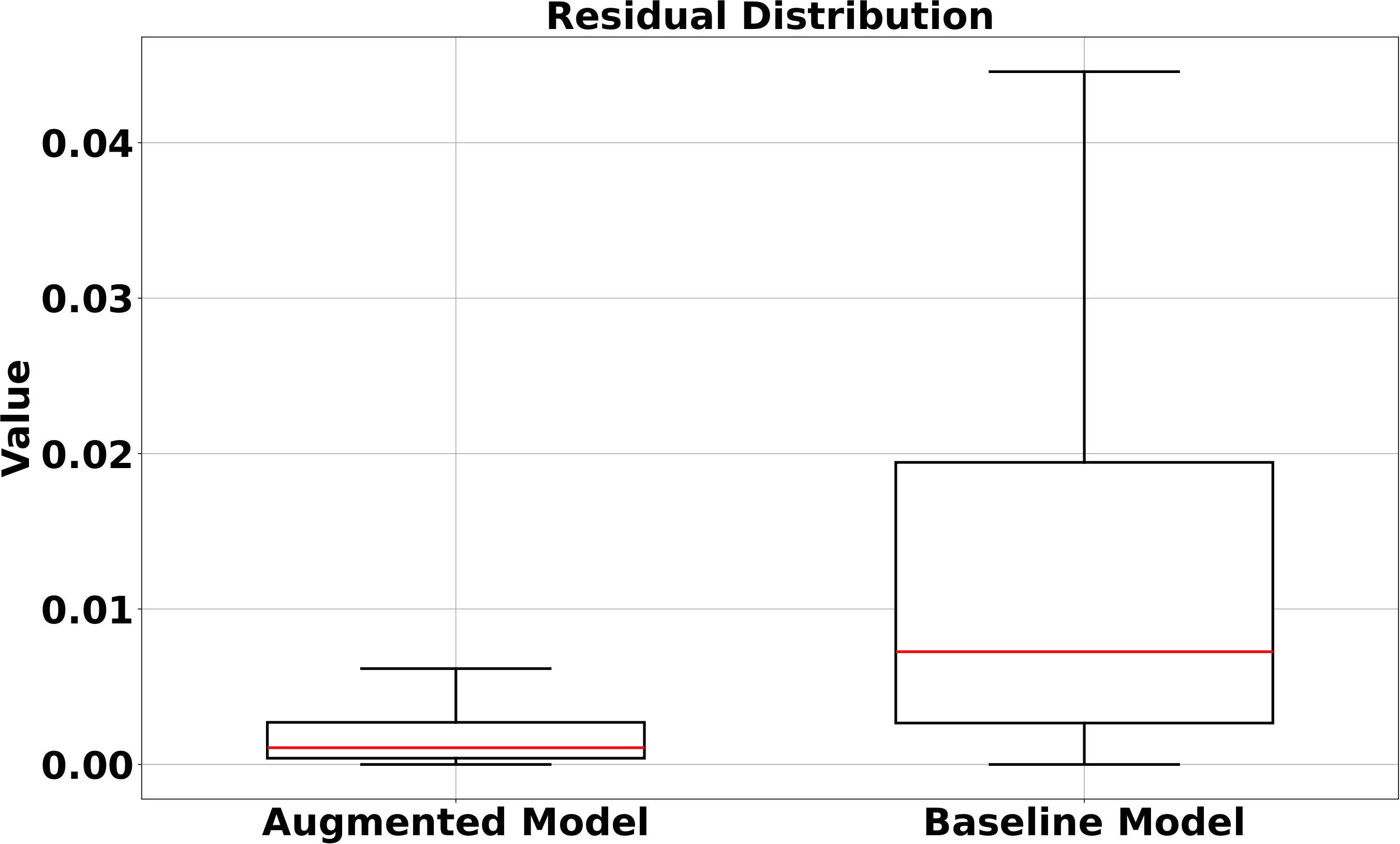}
    \caption{ \footnotesize{Box plot of minimum residuals between sampled worst-case obstacle clearances and ground truth. The augmented model achieves lower median error and reduced variability compared to the baseline, indicating sharper and more consistent alignment with ground-truth worst-case obstacle clearances.}}
    \label{fig:residual_box_plot}
    \vspace{-0.5cm}
\end{figure}

\begin{figure*}[t!]
    \centering
    \includegraphics[scale=0.21]{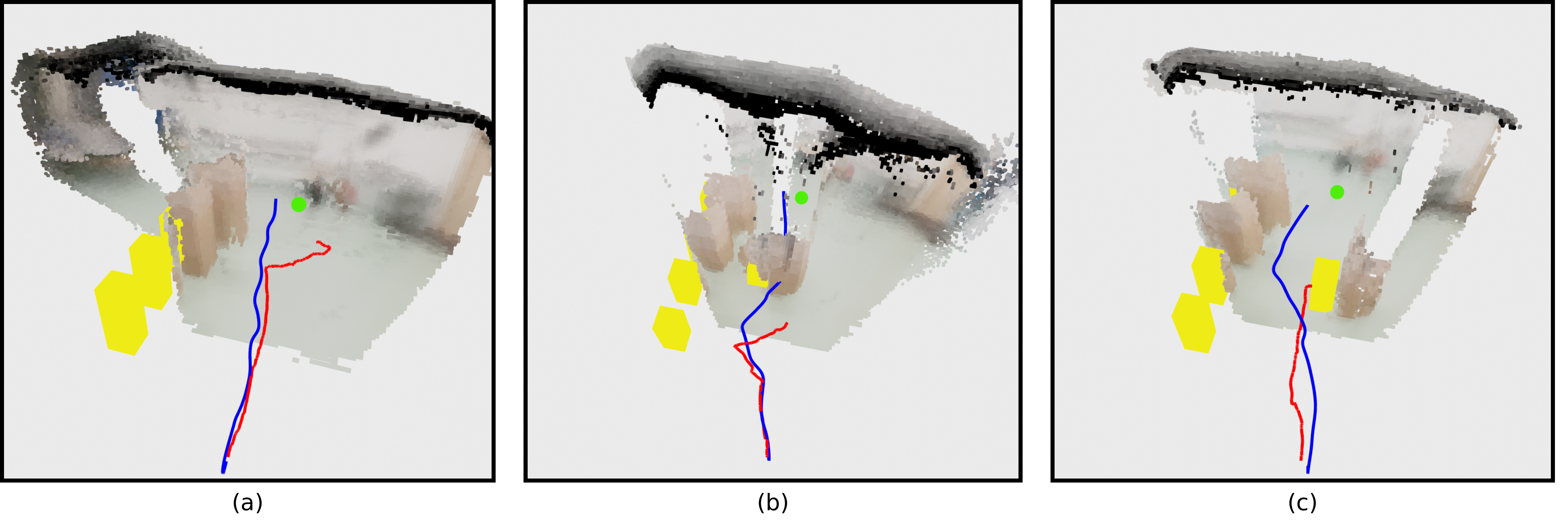}%
    \caption{\footnotesize{Comparison between our approach (blue) and MonoNav \cite{simon2023mononav} (red); goal in green. The noise in the estimated depth translates to erroneous 3D occupancy maps-the yellow cuboids (ground-truth obstacles) do not align with their reconstructed point clouds resulting in MonoNav getting stuck or colliding (b-c). In contrast, our approach is able to avoid the yellow cuboids based on the noisy estimated depth. We do not use MonoNav’s 3D reconstruction; trajectories are overlaid solely for visualization and comparison.}}
     \label{fig:mononav_comparison}
     \vspace{-0.3cm}
\end{figure*}

\subsection{Validation of the Learned Collision Model}
\noindent In this subsection, we validate the task-aware learning of the collision model (Fig.~\ref{fig:aug_nll_pipeline}) and compare with the baseline approach (Fig.~\ref{fig:vanilla_nll_pipeline}). 

\noindent \textbf{Predictive Accuracy:}We analyze residuals between sampled worst-case obstacle clearances and ground truth. For each test case, we record the minimum absolute residual between samples from the predictive distribution and the ground truth, indicating whether any sample captures the true worst-case clearance. As shown in Fig.~\ref{fig:residual_box_plot}, the task-aware augmented model of Fig.~\ref{fig:aug_nll_pipeline} yields lower median residuals and a tighter inter-quartile range than the baseline (Fig.~\ref{fig:vanilla_nll_pipeline}), indicating more consistent and reliable predictions.\\
\noindent \textbf{Distributional Metrics:} We evaluate three distributional metrics - Energy Score (ES), Probability Integral Transform (PIT), and Expected Calibration Error (ECE). The ES \cite{Gneiting01032007} measures both accuracy and sharpness, with lower values indicating predictions closer to the ground truth. Perfect calibration yields uniformly distributed PIT values \cite{repec:nbr:nberte:0215} on $[0,1]$, while deviations from uniformity are summarized by ECE \cite{Naeini2015ObtainingWC}, where $\text{ECE}\approx0$ denotes ideal calibration.

Additionally, to evaluate safety, we compute four outcomes on the test-set: \emph{Correctly Safe (CS)}, \emph{Correctly Unsafe (CU)}, \emph{Unsafe (U)}, and \emph{Conservative (C)}. We define safety-critical rates of Unsafe (UR) and Conservative (CR):
\[
\text{UR} = \text{U}/(\text{U} + \text{CU}) \times 100\%, \quad
\text{CR} =\text{C}/(\text{C} + \text{CS}) \times 100\%
\]
UR measures dangerous failures where unsafe states are overlooked, while CR captures over-conservatism where safe states are misclassified. For each outcome, risk $r_{MMD}^{emp}$ is computed from predicted clearances and compared against the ground-truth risk from $d_{gt}$.\\
\indent Table~\ref{table_results_metrics} shows that the $r_{MMD}^{emp}$-based augmented collision model (Fig.~\ref{fig:aug_nll_pipeline}) surpasses the baseline (Fig.~\ref{fig:vanilla_nll_pipeline}) across all metrics: UR $\downarrow$50\%, CR $\downarrow$80\%, ES $\downarrow$74\%, and ECE $\downarrow$20\%. These gains stem from task-aware variance supervision aligning uncertainty with collision risk, unlike the NLL-only baseline(Fig.~\ref{fig:vanilla_nll_pipeline}), which produces mis-calibrated and unreliable predictions. The same trend appears in navigation results (Table~\ref{table_results_ablation}), where the task-aware model reduces collisions by $4$ times. Table~\ref{table_results_ablation} also shows performance with deterministic collision model that performs the worst.

\subsection{Comparison with ROS Navigation Stack (ROSNAV)}
\noindent We benchmark our method against a classical planning stack from ROSNAV, which uses global planning followed by Dynamic Window Approach (DWA) for local control on costmaps built from estimated point clouds. To ensure a fair comparison with our planner's constraints (Alg.~\ref{algo_1}), we evaluate ROSNAV at two speed limits: 0.5 m/s and 1.0 m/s.
This classical approach fails mainly due to its inability to handle noise in the estimated depth. Offsets between estimated and true point clouds lead to false free space and collisions (Fig.~\ref{fig:teaser}(a)(3-4)), while noisy estimates inflate costmaps near goals, causing planning failures. These results show that such complex, dynamic errors cannot be corrected by simple heuristics.
Table~\ref{table_results_ros_vs_ours} quantifies these trends: ROSNAV often collides or fails at higher speeds, whereas our $r_{MMD}^{emp}$-based planner cuts collisions by up to 7 times and eliminates planning failures.

\begin{table}[!t]
\centering
\captionsetup{justification=centering}
\caption{\small{Validation Metrics for the Learned Collision Model}}
\vspace{2mm}
\renewcommand{\arraystretch}{1.5}
\resizebox{0.95\columnwidth}{!}{%
    \begin{tabular}{|>{\centering\arraybackslash}c|
                    >{\centering\arraybackslash}c|
                    >{\centering\arraybackslash}c|
                    >{\centering\arraybackslash}c|
                    >{\centering\arraybackslash}c|}
    \hline
    \textbf{Method} & \makecell[c]{\textbf{UR (\%)} \\ \textbf{(lower is better)}} & \makecell[c]{\textbf{CR (\%)} \\ \textbf{(lower is better)}} & \makecell[c]{\textbf{ES} \\ \textbf{(lower is better)}} & \makecell[c]{\textbf{ECE over PIT} \\ \textbf{(lower is better)}} \\
    \hline
    \makecell[c]{Alg.~\ref{algo_1} with $r_{MMD}^{emp}\text{-based}$ \\ Augmented Model (Fig.~\ref{fig:aug_nll_pipeline})} & \textbf{1.3} & \textbf{5.36} & \textbf{0.0236} & \textbf{0.0893} \\
    \hline
    \makecell[c]{Alg.~\ref{algo_1} with \\ Baseline Model (Fig.~\ref{fig:vanilla_nll_pipeline})} & 2.97 & 30.88 & 0.0917 & 0.1142 \\
    \hline
    \end{tabular}
}
\label{table_results_metrics}
\vspace{-0.2cm}
\end{table}

\begin{table}[!t]
\centering
\captionsetup{justification=centering}
\caption{\small{Navigation with and without Task-Aware Training}}
\vspace{2mm}
\renewcommand{\arraystretch}{1.5}
\resizebox{0.95\columnwidth}{!}{%
    \begin{tabular}{|>{\centering\arraybackslash}c|
                    >{\centering\arraybackslash}c|
                    >{\centering\arraybackslash}c|
                    >{\centering\arraybackslash}c|
                    >{\centering\arraybackslash}c|}
    \hline
    \textbf{Method} & \textbf{\% Collisions} & \textbf{\% Stuck} & \textbf{Avg. Speed (m/s)} & \textbf{Max. Speed (m/s)} \\
    \hline
    \makecell[c]{Alg.~\ref{algo_1} with $r_{MMD}^{emp}\text{-based}$ \\ Augmented Model (Fig.~\ref{fig:aug_nll_pipeline})}  & \textbf{6.6} & \textbf{0} & 0.32 & \textbf{1.02} \\
    \hline
    \makecell[c]{Alg.~\ref{algo_1} with \\ Baseline Model(Fig.~\ref{fig:vanilla_nll_pipeline})} & 28.3 & \textbf{0} & 0.35 & 0.94\\
     \hline
    \makecell[c]{Deteministic } & 41.5 & \textbf{0} & \textbf{0.52} & 1.01 \\
    \hline
    \end{tabular}
}
\label{table_results_ablation}
\vspace{-0.2cm}
\end{table}

\begin{table}[!t]
\centering
\captionsetup{justification=centering}
\caption{\small{Comparison with ROS Navigation Stack}}
\vspace{2mm}
\renewcommand{\arraystretch}{1.5}
\resizebox{0.95\columnwidth}{!}{%
    \begin{tabular}{|>{\centering\arraybackslash}c|
                    >{\centering\arraybackslash}c|
                    >{\centering\arraybackslash}c|
                    >{\centering\arraybackslash}c|
                    >{\centering\arraybackslash}c|}
    \hline
    \textbf{Method} & \textbf{\% Collisions} & \textbf{\% Stuck} & \textbf{Avg. Speed (m/s)} & \textbf{Max. Speed (m/s)} \\
    \hline
    \makecell[c]{Alg.~\ref{algo_1} with $r_{MMD}^{emp}\text{-based}$ \\ Augmented Model}  & \textbf{6.6} & \textbf{0} & 0.32 & 1.02 \\
    \hline
    \makecell[c]{ROSNAV \\ (0.5 m/s)} & 25 & 51.6 & 0.38 & 0.59 \\
    \hline
    \makecell[c]{ROSNAV \\ (1 m/s)} & 48.3 & 48.3 & \textbf{0.73} & \textbf{1.15} \\
    \hline
    \end{tabular}
}
\label{table_results_ros_vs_ours}
\vspace{-0.5cm}
\end{table}

\vspace{-0.3cm}
\subsection{Comparison with MonoNav \cite{simon2023mononav}}
\noindent To highlight the benefits of our uncertainty-aware framework, we benchmark it against MonoNav \cite{simon2023mononav}, a baseline that plans directly on 3D reconstructions from estimated depth. In a series of quadcopter navigation experiments (40 total) through increasingly cluttered 2.5D environments (Fig.~\ref{fig:mononav_comparison}), we demonstrate the critical failure modes of ignoring depth uncertainty.

As shown in Fig.~\ref{fig:mononav_comparison}, noise in MonoNav’s depth perception causes flawed reconstructions-tolerable in sparse scenes (a) but fatal in cluttered ones, leading to entrapment (b) or collision (c). Its simplistic primitive-based planner further limits adaptability to such errors. In contrast, our method explicitly incorporates depth uncertainty into risk-aware trajectory optimization, enabling successful navigation even in complex scenes. The executed trajectory (blue) consistently avoids all ground-truth obstacles (yellow cuboids), as confirmed by Table~\ref{table_results_mononav_vs_ours} and the accompanying video showing smoother, faster motions than MonoNav.

\begin{table}[!t]
\centering
\captionsetup{justification=centering}
\caption{\small{Comparison with MonoNav}}
\vspace{2mm}
\renewcommand{\arraystretch}{1.5}
\resizebox{0.95\columnwidth}{!}{%
    \begin{tabular}{|c|
                    c|c!{\color{black}\vrule width 1.5pt}c|c|}
    \hline
    \multirow{2}{*}{\textbf{Method}} & 
    \multicolumn{2}{c!{\color{black}\vrule width 1.5pt}}{\textbf{Easy Setting}} &
    \multicolumn{2}{c|}{\textbf{Moderately Difficult Setting}} \\
    \cline{2-5}
    & \textbf{\% Collisions} & \textbf{\% Stuck} & \textbf{\% Collisions} & \textbf{\% Stuck} \\
    \hline
    \makecell[c]{Alg.~\ref{algo_1} with $r_{MMD}^{emp}\text{-based}$ \\ Augmented Model} 
        & \textbf{0} & \textbf{0} & \textbf{0} & \textbf{0} \\
    \hline
    \makecell[c]{MonoNav \cite{simon2023mononav}} 
       & \textbf{0} & \textbf{0} & 20 & 80 \\
    \hline
    \end{tabular}
}
\label{table_results_mononav_vs_ours}
\vspace{-0.2cm}
\end{table}

\subsection{Comparison with End-to-End Model NoMaD\cite{10610665}}
\noindent NoMaD~\cite{10610665} is a foundational model for image-based goal-directed navigation. To focus on its obstacle avoidance capabilities, we evaluate it in simplified scenarios where the goal lies directly ahead and is visible from the start position, requiring only obstacle avoidance. We fine-tuned NoMaD using tele-operation data ($\sim$10k training points) and evaluated it in a test environment very similar to its training setup, giving it the best possible conditions to perform reliably.

Table~\ref{table_results_nomad_vs_ours} shows that NoMaD completes very few runs without collision, consistent with prior results~\cite{simon2023mononav}, which report over $50\%$ collision rates even in simpler hallway environments. Our settings were more complex, leading to even higher failure rates.

\begin{table}[!t]
\centering
\captionsetup{justification=centering}
\caption{\small{Comparison with NoMaD }}
\vspace{2mm}
\renewcommand{\arraystretch}{1.5}
\resizebox{0.95\columnwidth}{!}{%
    \begin{tabular}{|>{\centering\arraybackslash}c|
                    >{\centering\arraybackslash}c|
                    >{\centering\arraybackslash}c|
                    >{\centering\arraybackslash}c|
                    >{\centering\arraybackslash}c|}
    \hline
    \textbf{Method} & \textbf{\% Collisions} & \textbf{\% Stuck} & \textbf{Avg. Speed (m/s)} & \textbf{Max. Speed (m/s)} \\
    \hline
    \makecell[c]{Alg.~\ref{algo_1} with $r_{MMD}^{emp}\text{-based}$ \\ Augmented Model} & \textbf{8.3} & \textbf{0} & 0.28 & \textbf{1.05} \\
    \hline
    \makecell[c]{NoMaD \cite{10610665}} & 81.6 & \textbf{0} & \textbf{0.46} & 0.58\\
    \hline
    \end{tabular}
}
\label{table_results_nomad_vs_ours}
\vspace{-0.2cm}
\end{table}
\subsection{Computation Time}
\noindent On an RTX 3080 laptop, planning runs in $0.01$\,s with $0.02$\,s for parallel depth estimation. On a Jetson Orin, planning time is $0.065$\,s, thus demonstrating the feasibility of deploying our system on mobile robots with real-time re-planning.

\section{Conclusions and Future Work}
In this work, we showed that depth estimates from vision foundation models are unreliable for direct use in autonomous navigation. For example, occupancy maps built from them are error-prone, causing failures in cluttered scenes (e.g., ROS Navigation Stack, MonoNav\cite{simon2023mononav}). On the other hand, end-to-end vision-based methods such as NoMaD \cite{10610665} lack explicit constraint handling and also perform poorly. We address these limitations by formulating monocular vision-based navigation as a risk-aware planning problem over a learned probabilistic collision model. Our training pipeline enforces well-calibrated uncertainty, improving downstream risk estimation and planning performance. The proposed approach operates in real-time on both commodity laptops and Jetson Orin, with successful demonstrations on ground and aerial platforms. 

A limitation of our approach is the lack of temporal memory, as it ignores past images or point-clouds. This can cause oscillatory behavior in cluttered environments. Future work will focus on incorporating temporal information to mitigate this issue. Future work also includes extending our collision model to dynamic environments with time-varying collision models and accelerating planning through imitation learning. We also plan to explore end-to-end frameworks where collision model and action policy are jointly trained through demonstration data.

\bibliography{references}
\bibliographystyle{IEEEtran}

\end{document}